# Automatic Detection of Pulmonary Embolism using Computational Intelligence

Simon J. Scurell, Tshilidzi Marwala, and David Rubin

*Abstract*— This article describes the implementation of a system designed to automatically detect the presence of pulmonary embolism in lung scans. These images are firstly segmented, before alignment and feature extraction using PCA. In total 179 cases were collected from various medical institutions. Of the 179 cases, 125 (70%) were used for training (49 negative, 56 intermediate, 20 high), while 54 (30%) were used for validation (27 negative, 20 intermediate, 7 high). The network was trained using the Hybrid-Monte-Carlo method, resulting in a committee of 250 neural networks. The best system performance is achieved by using a network with 30 inputs, an image size of 64x64 and a VR of 90-90% which resulted in an AUC of 0.64. This compares favourably with previous work in which an AUC of 0.67 was achieved.

## I. INTRODUCTION

PULMONARY embolism (PE) is a potentially life-threatening condition, yet it is frequently treatable. The diagnosis of PE is based on a number of relatively non-specific findings including: presenting signs, shortness of breath, electrocardiographic findings, arterial blood gas abnormalities and chest X-ray changes [1]. In addition, a number of more specific imaging studies such as spiral computed tomography, pulmonary angiography and Ventilation/Perfusion (V/Q) scanning are employed. Obvious cases such as normal lungs and multiple large pulmonary emboli are usually easy to diagnose. The difficult diagnoses are those intermediate between the two extremes.

The accurate diagnosis of PE is a challenging problem, but one which radiographers make on a daily basis. The role of an automated system should at present not be thought of as a decision support system, and not as replacement to an experienced radiographer. The intrinsic characteristics of an automated tool would allow for a decrease in the inter-observer and intra-observer variability.

This paper is organized as follows, in Section II, a brief overview of VQ scanning is given. Section III deals with data collection. Sections IV and V deals with the implementation and training of the system. The results and conclusion in discussed in Sections VI and VII respectively.

## II. V/Q SCANNING

For a long time V/Q scanning has been the imaging protocol of choice for the evaluation of patients with suspected PE. The procedure has two parts: Perfusion and ventilation which enables visualization the blood and air flow in the lungs.

*1) Perfusion:* The patient is injected with 5mCi Technetium-99m MAA (macroaggregated albumin). This allows the radiologist to visualise the flow of blood in the patients lungs.

*2) Ventilation:* The patient is asked to inhale 20mCi of Technetium-99m DTPA (diethylenetriamine-pentaacetic acid) labelled aerosol which indicates airflow in the patients lungs. In some instances the patient may swallow the gas instead of inhaling it. Due to the deposition of the aerosol in the stomach and throat, artifacts in the image appear which need to be removed before diagnosis.

The scan consists of capturing images from six difference views, name the anterior, posterior, left lateral, right lateral, left posterior oblique and right posterior oblique.

## III. DATA COLLECTION

Retrospective General Hospital, Baragwanath Hospital and the Donald Gordon Medical Center. All of these institutions are located in the Johannesburg area in South Africa.

Each case consisted of 12 images, six for ventilation and six for perfusion which represent the views discussed earlier in Section II. All images were exported from APEX View in PCX file format. The images were then imported into MATLAB© and stored in a single structure. The V/Q scans where obtained using the following protocol:
- Ventilation: 20mCi Tc-99m DTPA
- Perfusion: 5mCi Tc-99m MAA

In total 179 cases where collected and studied retrospectively.

## IV. IMPLEMENTATION

There are several stages the images need to go through before a diagnosis can be made. Fig. 1 shows the block diagram for the system. Different sizes of lung images where tested, including 64x64, 32x32 and 16x16.

S. J. Scurell is with the School of Electrical and Information Engineering at the University of Witwatersrand, Johannesburg, South Africa; e-mail: s.scurell@ee.wits.ac.za).

Prof. T. Marwala is Carl Emily Fuchs Chair of Systems and Control at the School of Electrical and Information Engineering at the University of Witwatersrand, Johannesburg, South Africa; e-mail: t.marwala@ee.wits.ac.za ).

Prof. D. Rubin is professor at the School of Electrical and Information Engineering at the University of Witwatersrand, Johannesburg, South Africa; e-mail: d.rubin@ee.wits.ac.za ).

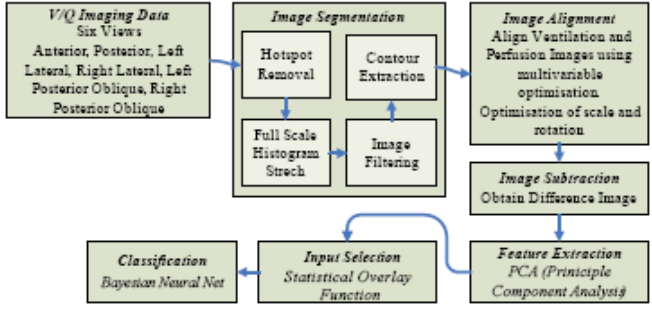

Fig. 1. System Block Diagram

*A. Image Segmentation*

All the lung images must at first be segmented. This removes unwanted background noise which may include areas of radiation in the stomach and trachea.

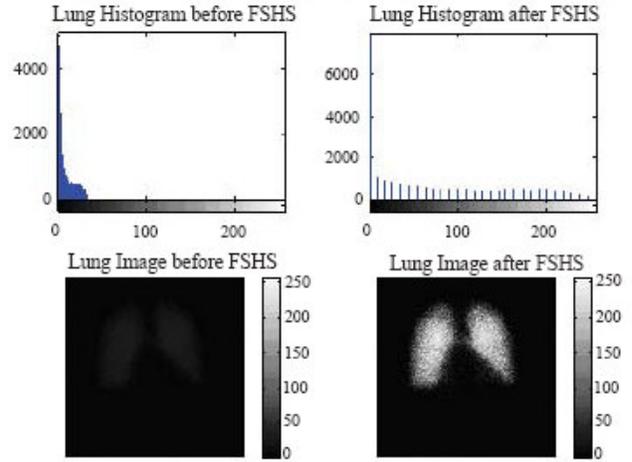

Fig. 2. Full Scale Histogram Stretch. The left images represent the images before the FSHS is performed, notice the poor contrast in the histogram and the lungs are hardly visible. The right images represent the images after a FSHS is performed showing an increase in contrast.

The registration of medical images can be achieved using several methods, such as using iterative processes that estimate affine transformations [2] between a reference image and a target image. There are two methods which described the use of neural networks [3], [4] in image registration. Multi-layer perceptron (MLP) methods have been used to find the edges of the lung from an MRI image [3]. The disadvantage of this method is it semi-automated, so would need human assistance. Another method describes the input images being represented by Gabor Wavelets to provide feature vectors [4].

Once the images have been registered, relevant features need to be extracted, which describe adequately the image information. Several methods for this are described in [5], such as The Karhunen-Loéve Transform, Principle Component Analysis, Discrete Time Fourier Transform (DFT), Hadamard Transform, Haar Transform and Discrete Time Wavelet Transform (DTWT).

*1) Hotspot Removal:* Hot spots are generally caused by the radiopharmaceutical getting trapped in a small area of the lung due to obstructive lung disease or the poor technical quality [6] of the radio nucleotides. Hot spot removal provides a reduction in areas of high intensity relative to the surrounding area in the image. The equation for determining a hotspot is given in (1) and is taken from [6]. Generally, hotspot removal is only applied to ventilation images.

$$HS = (x, y) \in C_v ; \frac{V(x, y) - \hat{V}}{\hat{s}} \geq q \quad (1)$$

*2) Full Scale Histogram Stretch:* A Full Scale Histogram Stretch (FSHS) is applied to each image; this greatly increases the contrast in the image and allows for more accurate contour detection. The equation for the FSHS is given below in (2).

$$V_o(x, y) = V_B(x, y) \times \left( \frac{K-1}{B-A} \right) \quad (2)$$

In this implementation K is set to 256, while *A* and *B* are the minimum and maximum image intensity respectively. Fig. 2 shows the effects of performing a FSHS on an anterior lung scan. The histogram in the top left is typical of an image exhibiting low contrast. This is confirmed in the image in Fig. 2 in which the lung is barely visible. The histogram on the right shows much better contrast, which can be seen in the lung image after the FSHS.

*3) Filtering and Contour Extraction:* The image is then filtered to reduce noise in the image and provide smoother contours. Iso-contours are calculated and the lung is segmented from the background. An example of an anterior lung image with overlayed iso-contours can be seen in Fig. 3, while Fig. 4 shows the lung images before and after segmentation. The views, from left to right, are anterior, posterior, left lateral, right lateral, left posterior oblique and right posterior oblique.

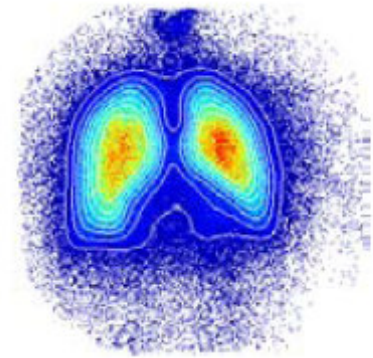

Fig. 3. Anterior lung with iso-contours overlayed.

*4) Dethroathing and Destomach:* Dethroating involves the removal of throat artifacts from the ventilation images. Throat artifacts, caused by the radio pharmaceutical getting trapped in the patients trachea, leads to areas of high image intensity.

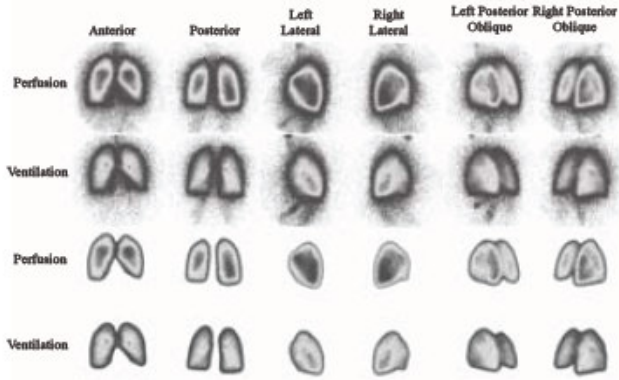

Fig. 4. Images before segmentation. From left to right the images are Anterior, Posterior, Left Lateral, Right Lateral, Left Posterior Oblique and Right Posterior oblique. Top row shows perfusion scans while the bottom row shows ventilation scans.

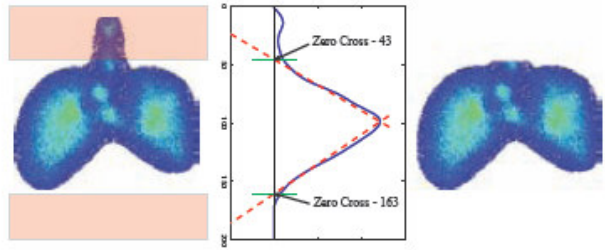

Fig. 5. Image dethroating. The left hand image shows the lung before dethroating. The right image shows the lungs after a dethroating is performed.

Stomach artifacts are caused by the patient swallowing some the radio pharmaceutical instead of inhaling it. The radio pharmaceutical then settles in the stomach and shows up on the image. These artifacts must be removed for the image alignment and classification to be most effective. The entire procedure is show in Fig. 5. The areas shaded in red indicate which areas are to be removed from the image.

### B. Image Alignment

As pulmonary embolism is identified by matched defects in perfusion and ventilation images, each VQ pair (there are six pairs) must be aligned before subtraction. More specifically areas where ventilation is present and perfusion absent are regarded as probable pulmonary emboli. To reduce the effects of lung defects on the alignment, the segmented images are converted to binary images. The alignment is accomplished using multi-variable (scale, rotation, x-translation, y-translation) genetic algorithms.

Fig. 6 shows the results of an image alignment algorithm using the GA with Schepp-Logan digital phantom images. The reference image represents the image that needs to be optimised, in other words the origin of the optimisation problem. The target image represents the destination of the optimisation. The 4 parameters, namely scale, rotation, x-translation and y-translation provide a transformation between the reference image and the target image. The transformation image represents the reference image, after it has been transformed with the optimized parameters. Table I shows a summary of the parameters found using the GA.

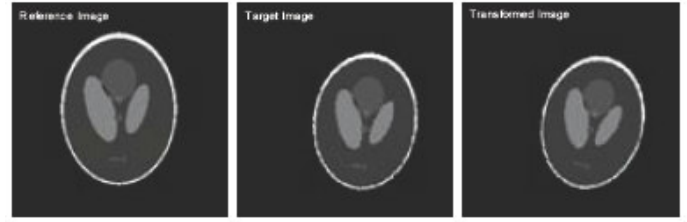

Fig. 6. Result of image alignment using Schepp-Logan digital phantoms. Summary of results can been found in Table I.

TABLE I
SUMMARY OF GA OPTIMISATION

| Parameter | Actual | Found | Error (%) |
|---|---|---|---|
| Scale | 0.9 | 0.88 | 2.3 |
| Rotation | 6.5º | 7.15º | 9.1 |
| X-translation | 25 | 27 | 7.4 |
| Y-translation | 15 | 16 | 6.25 |

### C. Image Subtraction

After the images all aligned the ventilation and perfusion images are subtracted. The algorithm subtracts the ventilation image from the perfusion image, areas with intensity values less than 0 indicate that there is more ventilation than perfusion in that specific area. The severity of the defect can then be quantified by taking a magnitude of pixel intensity in the subtraction image.

### D. Feature Extraction

PCA (principle component analysis) was performed on the images, from 16x16 to 64x64. As the image size gets smaller, for the same retained variability (VR), the number of required eigenvectors decreases. Conversely, for the same number of eigenvectors, the retained variability increases by approximately 10% for every half reduction in image size. This trend is most likely caused by a certain amount of variability being lost when reducing the image size.

### E. Feature Extraction

After the number of inputs has been reduced, it can be further minimised by using input selection methods. For this the statistical overlay function (SoF) is employed, taken from [7]. The SoF equation is given in (3).

$$\delta = \left\| \frac{\mu_1 - \mu_2}{\frac{\sigma_1 + \sigma_2}{2}} \right\| \qquad (3)$$

Each input to the system can be said to have a distribution of possible values. The goal of input selection is to choose those inputs whose distributions show the greatest amount of separation across the different classes of output. This concept is illustrated in Fig. 7. The greater the value of δ, the more separation there is between the two inputs. So δ between $\mu_1, \sigma_1$ and $\mu_3, \sigma_3$ would be greater than the δ between $\mu_1, \sigma_1$ and $\mu_2, \sigma_2$. The number of selected inputs was varied between 10, 20 and 30 for comparative purposes.

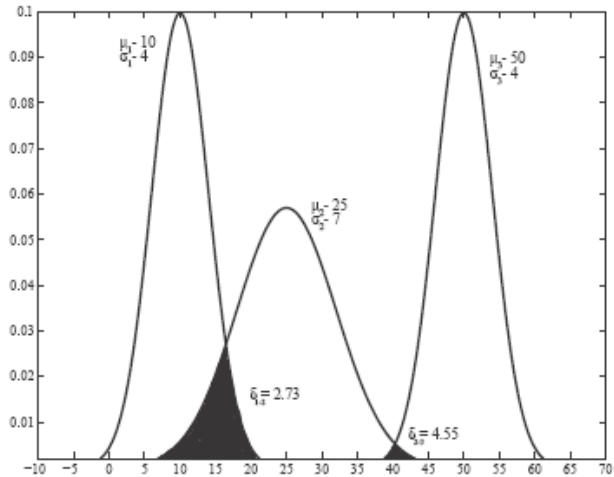

Fig. 7. Illustration of the Statistical Overlay Function (δ)

## V. SYSTEM TRAINING

The Bayesian MLP was trained in Matlab© using the NETLAB toolbox ([8]). The network consisted of a single input, hidden and output layer. The number of input nodes was varied between 10, 20 and 30. The hidden layer contained 5 nodes while the output layer contained 1 node.

Of the 179 cases, 125 (70%) were used for training (49 negative, 56 intermediate, 20 high), while 54 (30%) were used for validation (27 negative, 20 intermediate, 7 high). The network was trained using the Hybrid-Monte-Carlo method, resulting in a committee of 250 neural networks.

The scaled PCA inputs were fed into each of the 250 networks and the final classification was based upon the mean value of the committee output. Because each member of the committee gives an individual output, 95% confidence levels can be calculated by taking into account the standard deviations (σ) of the output distributions. So outputs that exhibit a small σ imply a high confidence, while outputs exhibiting a large σ imply a low confidence.

## VI. MAIN RESULTS

The VR, chosen during the PCA analysis is a parameter which was varied. A steep increase in training performance is gained between a VR of 70% and 75%. There also appears to be a gradual increase in validation performance with increasing VR. Validation performance also increased with input size.

Overall the negative cases are shown to be diagnosed with the highest degree of accuracy, followed by the high probability cases and lastly the intermediate probability cases. This would confirm what was mentioned earlier in that the intermediate cases are the most difficult to diagnose. The validation performance increases with image size, and the number of inputs to the neural network. Table II shows the sensitivities, specificities, positive and negative predictive values for the different output. The specificity and NPV show the highest means, indicating that PE can be reasonably excluded if the prediction is negative.

TABLE II
TABLE SHOWING SENSITIVITY, SPECIFICITY, POSITIVE PREDICTIVE VALUE (PPV) AND NEGATIVE PREDICTIVE VALUE (NPV) USING THE VALIDATION DATA SET

|  |  | High | Intermediate | Negative |
|---|---|---|---|---|
| **Sensitivity [%]** | max | 71.43 | 55.00 | 70.37 |
|  | min | 14.29 | 0.00 | 18.52 |
|  | mean | 46.83 | 30.84 | 48.35 |
| **Specificity [%]** | max | 95.92 | 94.44 | 77.14 |
|  | min | 68.11 | 68.00 | 61.37 |
|  | mean | 81.03 | 82.01 | 69.01 |
| **PPV [%]** | max | 71.43 | 62.50 | 61.54 |
|  | min | 6.25 | 0.00 | 33.33 |
|  | mean | 25.95 | 41.24 | 50.97 |
| **NPV [%]** | max | 95.92 | 79.07 | 77.14 |
|  | min | 88.68 | 62.96 | 55.10 |
|  | mean | 92.71 | 71.44 | 66.42 |

The best system performance is achieved by using a network with 30 inputs, an image size of 64x64 and a VR of 90-90%.

In order to compare these results with those of previous authors, a receiver operating characteristic (ROC) curve is needed. As most of the previous works only considered a 2-class classifier, it was decided that for comparative purposes the intermediate probability cases would be grouped together with the high probability cases to form a new "positive" class. The ROC curve, shown in Fig. 8, was generated from a network using an image size of 64x64, 30 inputs and a VR of 95%.

The area under curve (AUC) is 0.64. This compares to an AUC of 0.86 achieved in [9]. An interesting comparison is to the work done in [6] where an AUC of 0.85 was achieved when the gold standard was angiography, while an AUC of 0.67 was achieved when the gold standard was the consensus opinion of nuclear medicine physicians. This compares favourably with the AUC of 0.64 in this study, where the gold standard was also the opinion of a nuclear medicine physician.

## VII. CONCLUSION

Although performance is below that of previous works, it has been shown that it is feasible to diagnose PE automatically using a Bayesian Neural Network. The most critical areas in the process are those of image segmentation, feature extraction and input normalisation. Without correct scaling of the input data, the network simply does not "work". The image segmentation routines proved to be very effective in most circumstances. There were however a few instances where manual touch-up was needed. This was mostly done on ventilation images where the scans contained large amounts of noise.

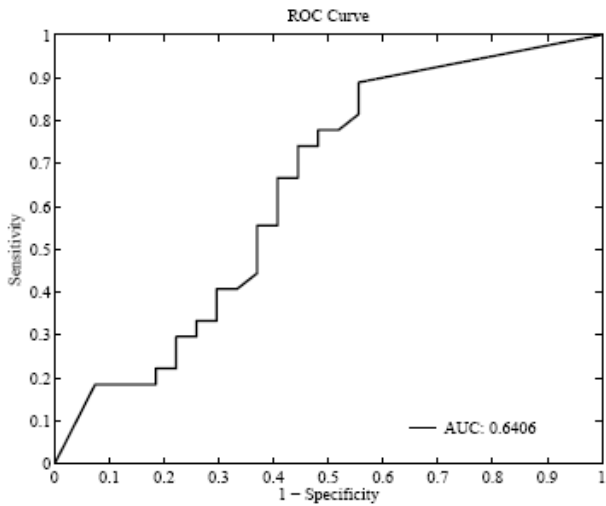

Fig. 8. ROC curve for Bayesian Classifier. 64x64 image size, using 30 inputs and a VR of 95\%. The AUC is 0.64. Intermediate and high probability cases are grouped together.

From the results, it would appear that the best system performance is achieved by having a network with 30 inputs using an image size of 64x64 and a VR of 90-95%. Intermediate cases were shown to be the most difficult to diagnose correctly, while specificity calculations show that PE can reasonably be excluded if the prediction is not positive.

This research could be furthered by the collection of more cases to provide a more balanced data set including equal numbers of each class for both training and validation. It is likely that a large increase in available cases will greatly improve system performance.

If possible, using a different *Gold Standard*, angiography for instance, may improve the performance of the Bayesian classifier.

ACKNOWLEDGMENT

I would like to thank the staff of the Chris Hani Baragwanath Hospital, Johannesburg General Hospital and the Donald Gordon Medical Centre for their assistance in obtaining the imaging data. A special thanks must go to Dr Carlos Liebhabe.

This work was supported by DENEL and the Ledger Project.

REFERENCES

[1] M. Rodger and P. Wells, "Diagnosis of pulmonary embolism," *Thrombosis Research*, vol. 103, pp. 224–238, 200.
[2] K. Cios, L. Goodenday, and J. Sacha, "Bayesian learning for cardiac SPECT image interpretation," *Artificial Intelligence in Medicine*, vol. 26, no. 1-2, pp. 109–143, September-October 2002.
[3] R. Damper and I. Middleton, "Segmentation of magnetic resonance images using a combination of neural networks and active contour models," *Medical Engineering & Physics*, vol. 26, no. 1, pp. 71–86, January 2003.
[4] G. Coppini, S. Diciotti, and G. Vialli, "Matching of medical images by self-organizing neural networks," *Pattern Recognition Letters*, vol. 25, no. 3, pp. 341–352, February 2004.
[5] K. Koutroumbas and S. Theodoridis, *Pattern Recognition*. Academic Press, 1999.
[6] A. Frigyesi, "An automated method for the detection of pulmonary embolism in V/Q scans," *Medical Image Analysis*, vol. 7, no. 7, pp. 341–349, 2003.
[7] T. Marwala, "Fault identification using neural networks and vibration data," Ph.D. dissertation, University of Cambridge, 2001.
[8] I. T. Nabney, "NETLAB," www, June 2002.
[9] A. Ericsson, A. Huart, A. Ekefjrd, K. strm, H. Holst, E. Evander, P. Wollmer, and L. Edenbrandt, "Automated interpretation of ventilation-perfusion lung scintigrams for the diagnosis of pulmonary embolism using Support Vector Machines," in *13th Scandinavian Conference on Image Analysis*. SCIA, June 2003.